\documentclass{article}

\usepackage{times}
\usepackage{graphicx} 

\graphicspath{ {figures/} }
\usepackage{amssymb}

\usepackage{amsmath}

\newcommand{\mb}{\mathbf}

\newcommand{\argmax}{\operatornamewithlimits{argmax}}

\newcommand{\Exp}[2]{\mathbb{E}_{#1}\left[#2\right]}

\usepackage{pdflscape}

\usepackage{algorithm}
\usepackage{algorithmic}

\usepackage{hyperref}

\usepackage[numbers]{natbib}

\usepackage[accepted]{icml2015}

\icmltitlerunning{RNN VAE Equivalence}

\begin{document}

\onecolumn

\icmltitle{
Note on Equivalence Between Recurrent Neural Network\\Time Series Models and Variational Bayesian Models
}

\icmlauthor{Jascha Sohl-Dickstein}{jascha@stanford.edu}
\icmlauthor{Diederik P. Kingma}{dpkingma@uva.nl}
\icmladdress{}


\vskip 0.3in


\begin{abstract}
We observe that the standard log likelihood training objective for a Recurrent Neural Network (RNN) 
model of time series data is equivalent to a variational Bayesian training objective, 
given the proper choice of generative and inference models. 
This perspective may motivate extensions to both RNNs and variational Bayesian models. 
We propose one such extension, where multiple particles are used for the hidden state of an RNN, 
allowing a natural representation of uncertainty or multimodality.
\end{abstract}

\section{Recurren Neural Networks (RNNs)}

\subsection{RNN definition}\label{sec rnn def} 

A Recurrent Neural Network (RNN) \cite{Haykin1999} has a visible state $\mb x^t$ at each time step, and a corresponding hidden state $\mb h^t$. The dynamics can be described in terms of two distributions $p\left( \mb h^t | \mb h^{t-1}, \mb x^{t-1} \right)$ and $p\left( \mb x^t | \mb h^{t} \right)$. Typically the state of the hidden units $\mb h^t$ is deterministic given $\mb x^{t-1}$ and $\mb h^{t-1}$, the values at the previous timestep, such that $\mb h^t = f\left( \mb h^{t-1}, \mb x^{t-1} \right)$. Taking slight liberties with notation, we indicate the distribution of the hidden units given their parents as:
\begin{align}
p\left( \mb h^t | \mb h^{t-1}, \mb x^{t-1} \right) & = \delta\left( \mb h^t - f\left( \mb h^{t-1}, \mb x^{t-1} \right) \right), \\
p\left( \mb H | \mb X \right) & = \delta\left( \mb H - F\left( \mb X \right) \right) 
.
\end{align}
Where $\mb X = \{\mb x^1 \cdots \mb x^T\}$ and $\mb H = \{\mb h^2 \cdots \mb h^T\}$ are the full trajectories over visible and hidden units. 
The initial hidden vector $\mb h^1$ is included as a model parameter.

\subsection{RNN training}

Training is typically performed by maximizing the log likelihood of this model, computed from a data distribution over trajectories $q\left( \mb X \right)$. Note that this empirical data distribution $q\left(\mb X \right)$ is most often simply a collection of datapoints, i.e. Dirac delta peaks.
\begin{align}
\hat{f}\left( \mb h^{t-1}, \mb x^{t-1} \right), \hat{p}\left( \mb x^t | \mb h^{t} \right) 
&  = \argmax_{
{f}\left( \mb h^{t-1}, \mb x^{t-1} \right), {p}\left( \mb x^t | \mb h^{t} \right)
}
\Exp{q\left( \mb X \right)}{ \log p\left( \mb X \right)}
\\
&  = \argmax_{
{f}\left( \mb h^{t-1}, \mb x^{t-1} \right), {p}\left( \mb x^t | \mb h^{t} \right)
}
\Exp{ q\left( \mb X \right)}{\Exp{p\left( \mb H | \mb X \right) }{ \sum_t \log p\left( \mb x^t | \mb h^{t} \right)}} \label{eq log L p exp}
\\
&  = \argmax_{
{f}\left( \mb h^{t-1}, \mb x^{t-1} \right), {p}\left( \mb x^t | \mb h^{t} \right)
}
\Exp{ q\left( \mb X \right)}{\Exp{ \delta\left( \mb H - F\left(  \mb X \right) \right)}{ \sum_t \log p\left( \mb x^t | \mb h^{t} \right)}}
\label{eq rnn train}
.
\end{align}

\section{Variational Bayesian Perspective}

Variational Bayesian methods where both the generative and inference models are trained against each other have recently proven very powerful for building probabilistic models of arbitrary distributions \cite{sminchisescu2006learning,Kingma2013,Gregor2013,Rezende2014,Ozair2014,SohlDickstein2015,Gregor2015}. Here we show how an RNN can be interpreted using a variational Bayesian framework.

\subsection{Inference model}

We take $p\left( \mb X, \mb H \right)$ from Section \ref{sec rnn def} to be the `generative' model, and introduce an `inference' model $q\left( \mb H | \mb X \right)$. We set the inference model to be identical to the corresponding conditional distribution in the generative model,
\begin{align}
q\left( \mb h^t | \mb x^{t-1}, \mb h^{t-1} \right) &= p\left( \mb h^t | \mb x^{t-1}, \mb h^{t-1} \right) \label{eq inference}
\\
q\left( \mb H | \mb X \right) 
&= \prod_t q\left( \mb h^t | \mb x^{t-1},  \mb h^{t-1} \right) 
\\
&= p\left( \mb H | \mb X \right) \label{eq p q equiv}
.
\end{align}

\subsection{Log likelihood bound} \label{sec log bound}

We now derive a variational bound $K$ on the data log-likelihood $L = \Exp{q\left( \mb X \right)}{\log p(\mb X)}$ for these generative and inference distributions,
\begin{align}
L &= \Exp{q\left( \mb X \right)}{\log p(\mb X)}
\\
&= \Exp{ q\left( \mb X \right)}{\log \int d\mb H\ p\left( \mb H, \mb X \right)}
\\
&= \Exp{ q\left( \mb X \right)}{\log \Exp{q\left( \mb H | \mb X \right)}{\frac{
    p\left( \mb H, \mb X \right) 
    }{
    q\left( \mb H | \mb X \right) }}}
\\
&\geq
\Exp{ q\left( \mb X \right)}{
    \Exp{ q\left( \mb H | \mb X \right)}{ \log \frac{
    p\left( \mb H, \mb X \right) 
    }{
    q\left( \mb H | \mb X \right) 
    }}} \tag*{(Jensen's inequality)}
\\
 = K &=
\Exp{q\left( \mb X \right)}{\Exp{ q\left( \mb H | \mb X \right)}{ \log \frac{
    \prod_t p\left( \mb h^t | \mb x^{t-1}, \mb h^{t-1} \right) p\left( \mb x^t | \mb h^t \right)
    }{
    \prod_t q\left( \mb h^t | \mb x^{t-1}, \mb h^{t-1} \right) 
    }}} \\
&=
\Exp{q\left( \mb X \right)}{\Exp{ q\left( \mb H | \mb X \right)}{ \log \frac{
    \prod_t p\left( \mb h^t | \mb x^{t-1}, \mb h^{t-1} \right) p\left( \mb x^t | \mb h^t \right)
    }{
    \prod_t p\left( \mb h^t | \mb x^{t-1}, \mb h^{t-1} \right) 
    } }}\\
&=
\Exp{ q\left( \mb X \right)}{\Exp{ q\left( \mb H | \mb X \right)}{ \sum_t \log 
    p\left( \mb x^t | \mb h^t \right)}} \label{eq noisy}
    \\
&=
\Exp{ q\left( \mb X \right)}{\Exp{ \delta\left( \mb H - F\left( \mb X \right) \right)}{ \sum_t \log 
    p\left( \mb x^t | \mb h^t \right)}} \label{eq equiv obj}
.
\end{align}

The bound in Equation \ref{eq equiv obj} is identical to the RNN training objective in Equation \ref{eq rnn train}. 
Therefore, for the choice of inference model in Equation \ref{eq inference}, the variational Bayesian training objective 
is identical to the standard log likelihood training objective.

\subsection{Optimality of noise-free hidden dynamics}
Often, noise-free dynamics is optimal w.r.t. $K$.
If the latent dynamics $p(\mb h^t|\mb h^{t-1}, \mb x^{t-1})$ is in the location-scale family with location $\mb \mu(\mb h^{t-1}, \mb x^{t-1})$ and scale $\mb \sigma$, we can parameterize the latent variables as $\mb h^t = f(\mb h^{t-1}, \mb x^{t-1}) + \mb \sigma \cdot \mb \epsilon^t$, where $f$ is a deterministic function and $\mb \epsilon^t \sim p(\mb \epsilon)$ is independent zero-centered noise per timestep. 
 Equation~\eqref{eq noisy} can be written in this so-called non-centered form~\cite{kingma2014efficient} as:
\begin{align}
\Exp{ q\left( \mb X \right)}{\Exp{ p(\mathbf{\epsilon})}{ \sum_t \log 
    p\left( \mb x^t | \mb h^t \right)}}
\end{align}
The inserted noise $\mb \epsilon^t$ has the effect of removing information about previous states from the hidden state $\mb h^t$, such that $\mb x^t$ will be harder to predict. This contribution of the noise to $\mb h^t$ can trivially be minimized by letting $\sigma \to 0$, i.e. by choosing $q(\mb H | \mb X) = \delta( \mb H - F( \mb X ) )$ (eq.~\eqref{eq equiv obj}).

\section{Discussion}

From one perspective it is a trivial observation that if $q\left(\mb H|\mb X\right) = p\left(\mb H|\mb X\right)$, 
then the variational Bayesian objective becomes the true log likelihood objective. From another perspective, 
it is non-obvious and interesting that due to its causal structure, a recurrent neural network can be viewed simultaneously 
as an inference and generative model, and that the inference model can be made trivially identical to the posterior 
of the generative model.

Note that the equivalence $q\left( \mb H | \mb X \right) = p\left( \mb H | \mb X \right)$ in Equation \ref{eq p q equiv} 
relies on the true posterior distribution $p\left(\mb H| \mb X\right)$ having the causal, factorial, structure
$p\left(\mb H| \mb X\right) = \prod_t p\left( \mb h^t | \mb x^{t-1},  \mb h^{t-1} \right)$. 
This structure stems from the 
deterministic dynamics of the RNN, and would not hold in general if $p\left( \mb h^t | \mb h^{t-1}, \mb x^{t-1} \right)$ were not a delta function. 
In this case the variational bound on the log likelihood 
in Equation \ref{eq noisy} 
would continue to hold, 
but it would no longer be identical to the true log likelihood in Equation \ref{eq log L p exp}.

This perspective on RNNs as consisting of matching inference and generative models may suggest 
natural extensions to the RNN framework, or novel model forms for variational Bayesian methods. 
As one example, it suggests the use of multiple inference particles in an RNN, 
which may allow 
more complex and multimodal distributions over visible units 
to be captured by 
simpler and lower dimensional hidden representations.

%
%

\subsection{Multiple particles}

RNNs are often called upon to represent a multimodal distribution over $p\left( \mb x^t | \mb x^1 \cdots \mb x^{t-1} \right)$ (for instance, a distribution over words in the context of language models). Since the hidden state of an RNN is deterministic, it must capture this multimodal distribution using a single high dimensional vector $\mb h^t$.

In variational inference, a multimodal posterior can be approximated using multiple samples from the inference model. This raises the possibility of training an RNN with a multimodal distribution over hidden units. 
This has the potential to reduce the required complexity of the RNN. Rather than forcing a high dimensional unimodal distribution over hidden units to represent a multimodal distribution over visible units, instead multiple modes in a lower dimensional hidden representation can be made to correspond to the multiple modes over the visible units.

Specifically, the hidden state can be extended to consist of $L$ samples $\mb H = \{ \mb H_1, \mb H_2, \cdots, \mb H_L\}$. As shown in Appendix \ref{app mult part} these multiple samples can be averaged over in the variational Bayesian framework. This leads to the following modification of the training objective from Equation \ref{eq equiv obj},
\begin{align}
L \geq K &=
\Exp{ q\left( \mb X \right)}{\Exp{ \delta\left( \mb H - F\left( \mb X \right) \right)}{ \sum_t \log \left(
\frac{1}{L} \sum_{l=1}^L 
    p\left( \mb x^t | \mb h_l^t \right)
\right)
}} 
.
\label{eq K multi}
\end{align}
The multiple samples can be initialized with different (learned) initial vectors $\mb h^1_l$, allowing them to explore different modes despite being governed by the same deterministic dynamics.

\newpage
\appendix

\normalsize

\part*{Appendix}

\section{Multiple particles in variational Bayesian models}\label{app mult part}

Here we modify the derivation in Section \ref{sec log bound} to include multiple particles, leading to Equation \ref{eq K multi}.
\begin{align}
p\left(\mb X\right) &= \frac{1}{L} \sum_{l=1}^L \int d\mb H_l\ p\left( \mb H_l, \mb X \right)
\\
 &= \frac{1}{L} \sum_{l=1}^L \int d\mb H_l\  \frac{
	p\left( \mb H_l, \mb X \right) }{
	q\left( \mb H_l | \mb X \right)    }
	q\left( \mb H_l | \mb X \right)
\\
 &= \frac{1}{L} \sum_{l=1}^L \int d\mb H_l\  \frac{
	p\left( \mb H_l, \mb X \right) }{
	q\left( \mb H_l | \mb X \right)    }
	q\left( \mb H_l | \mb X \right) 
	\prod_{l' \neq l}
	\int d\mb H_{l'}\ q\left( \mb H_{l'} | \mb X \right)
\\
 &= \frac{1}{L} \sum_{l=1}^L \int d\mb H\  \frac{
	p\left( \mb H_l, \mb X \right) }{
	q\left( \mb H_l | \mb X \right)    }
	q\left( \mb H_l | \mb X \right) 
	\prod_{l' \neq l}
	q\left( \mb H_{l'} | \mb X \right)
\\
q\left( \mb H | \mb X \right) &=
\prod_l q\left( \mb H_l | \mb X \right) \\
p\left(\mb X\right)	
 &= \frac{1}{L} \sum_{l=1}^L \int d\mb H\ 
 		q\left( \mb H | \mb X \right) 
  \frac{
	p\left( \mb H_l, \mb X \right) }{
	q\left( \mb H_l | \mb X \right)    }
\\
 &= \int d\mb H\ 
 		q\left( \mb H | \mb X \right) 
\left( \frac{1}{L} \sum_{l=1}^L
  \frac{
	p\left( \mb H_l, \mb X \right) }{
	q\left( \mb H_l | \mb X \right)    }
\right)
\\
L &= \Exp{q\left( \mb X \right)}{\log p(\mb X)}
\\
&= \Exp{ q\left( \mb X \right)}{\log \Exp{q\left( \mb H | \mb X \right)}
{
\frac{1}{L} \sum_{l=1}^L
  \frac{
	p\left( \mb H_l, \mb X \right) }{
	q\left( \mb H_l | \mb X \right)    }
	}} \label{eq app L}
\\
L &\geq
\Exp{ q\left( \mb X \right)}{\Exp{ \delta\left( \mb H - F\left( \mb X \right) \right)}{ \sum_t \log \left(
\frac{1}{L} \sum_{l=1}^L 
    p\left( \mb x^t | \mb h_l^t \right)
\right)
}} \label{eq app final}
,
\end{align}
where the steps between Equations \ref{eq app L} and \ref{eq app final} parallel those in Section \ref{sec log bound}.

\bibliography{rnn_varauto}
\bibliographystyle{plain}

\end{document}